\documentclass{article}

\usepackage{microtype}
\usepackage{graphicx}
\usepackage{subfigure}
\usepackage{booktabs} 

\usepackage{hyperref}

\usepackage[preprint]{icml2024/icml2024}

\usepackage{multirow}

\usepackage{amsmath}
\usepackage{amssymb}
\usepackage{mathtools}
\usepackage{amsthm}

\usepackage{ulem}

\usepackage{color}

\usepackage{graphicx}
\usepackage{caption}

\usepackage[capitalize,noabbrev]{cleveref}

\theoremstyle{plain}
\newtheorem{theorem}{Theorem}[section]

\theoremstyle{definition}
\newtheorem{definition}[theorem]{Definition}

\theoremstyle{remark}

\usepackage[textsize=tiny]{todonotes}

\usepackage{multirow}

\begin{document}

\twocolumn[
\icmltitle{Data-free Weight Compress and Denoise for Large Language Models
}




\begin{icmlauthorlist}
\icmlauthor{Runyu Peng}{Fudan}
\icmlauthor{Yunhua Zhou}{Pujiang}
\icmlauthor{Qipeng Guo}{Pujiang}
\icmlauthor{Yang Gao}{Pujiang}
\icmlauthor{Hang Yan}{Pujiang}
\icmlauthor{Xipeng Qiu}{Fudan}
\icmlauthor{Dahua Lin}{Pujiang}

\end{icmlauthorlist}

\icmlaffiliation{Fudan}{School of Computer Science, Fudan University, Shanghai, China}
\icmlaffiliation{Pujiang}{Shanghai AI Lab, Shanghai, China}


\icmlcorrespondingauthor{Runyu Peng}{22210240041@fudan.edu.cn}


\vskip 0.3in
]




\begin{abstract}

Large Language Models (LLMs) are reshaping the research landscape in artificial intelligence, particularly as model parameters scale up significantly, unlocking remarkable capabilities across various domains. Nevertheless, the scalability of model parameters faces constraints due to limitations in GPU memory and computational speed.
To address these constraints, various weight compression methods have emerged, such as Pruning and Quantization. Given the low-rank nature of weight matrices in language models, the reduction of weights through matrix decomposition undoubtedly holds significant potential and promise.
In this paper, drawing upon the intrinsic structure of LLMs, we propose a novel approach termed~\textbf{Data-free Joint Rank-k Approximation} for compressing the parameter matrices. Significantly, our method is characterized by without necessitating additional involvement of any corpus, while simultaneously preserving orthogonality in conjunction with pruning and quantization methods. We achieve a model pruning of 80\% parameters while retaining 93.43\% of the original performance without any calibration data. Additionally, we explore the fundamental properties of the weight matrix of LLMs undergone Rank-k Approximation and conduct comprehensive experiments to elucidate our hypothesis.

\end{abstract}

\section{Introduction}

The ascendancy of large language models (LLMs) is reshaping the research paradigm of natural language processing (NLP). The prevalence of high-quality open-source models, such as LLaMA~\cite{touvron2023llama, touvron2023llama2} and Mistral-7B~\cite{jiang2023mistral}, is exacerbating the trend of researchers from various fields being engaged. Particularly, with the scaling of language models, the emerging capabilities empower language models to be generalized across different fields with appropriate prompts. Nevertheless, scaling language models poses considerable challenges. In addition to the substantial hardware costs involved, the training of large language models necessitates complex distributed systems due to memory constraints.

To break through the bottlenecks posed by GPU memory and computational speed, a variety of techniques have emerged, with Quantization~\cite{park2018value, dettmers2023qlora, lin2023awq, shen2019qbert, yao2022zeroquant} and Pruning~ \cite{jha2023train, xia2023sheared, ma2023llmpruner, zhang2023loraprune, sun2023simple, frantar2023sparsegpt} being representatives, extensively researched for their commendable performance. These methods typically employ numerical approximations and other strategies to operate the target LLMs with minimal computational cost. However, to attain comparable performance, on the one hand, these techniques necessitate fine-tuning for calibration, and on the other hand, the ensuing fine-tuning may introduce bias towards the distribution of the additional datasets, thereby compromising the model's generalization capability.

\textbf{Is there an effective approach to compress model parameters without the need for additional data?} Let us delve into this step by step. The weight parameters of existing language models are organized and contribute to fundamental operations in a matrix manner, which naturally inspires us to introduce matrix approximation for compressing model parameters. Notably, parameter compression based on matrix approximation is inherently well-grounded in a data-free context.


Furthermore, despite modifications for various purposes, existing language models share a common inductive bias in structure, all grounded in Transformer block stacks. Upon closer look to the essential components within the basic Transformer block, whether the attention module or the Feed-forward module, they can be construed as operators for matrix multiplication. The mapping space of these operators is defined by the matrix parameters within them. Existing approaches often approximate a specific matrix or independently approximate all matrices~\cite{lv-etal-2023-lightformer, sharma2023truth}, leading to discrepancies between the approximated mapping space and the original mapping space in terms of integrity. Therefore, this paper proposes a novel method - Joint Rank-k Approximation, which decomposes the matrices within the operator as a whole, aiming to reconstruct the mapping space of the original operator as closely as possible while compressing parameter weights.


Finally, utilizing widely accepted language model structures as an archetype, we not only demonstrate the superiority of Joint Rank-k approximation over existing matrix approximation methods but also further theoretically characterize its potential to enhance model performance and robustness. This is accomplished by effectively filtering out noise from high-order parameter matrices in low-intensity components of the spectral domain. In conjunction with theoretical analysis, we conducted comprehensive experiments to validate the effectiveness and generalization capabilities of our proposed method.
On down-stream tasks, compressed model retains 98.80 \% of original performance at a prune rate of 10\% and 93.43 \% at a prune rate of 20\%.
In summary, this paper introduces a foundational and versatile method, offering novel insights that may serve as inspiration for subsequent research in this field.

The contribution of this paper can be summarized as follows:
\begin{itemize}
\item This paper introduces a novel and efficient approach for compressing the weight parameters of language models, termed as Data-free Joint Rank-k Approximation. This method proves parameter-effective
while striving to preserve the consistency of the Transformer block mapping space. 
\item We deduce that by keeping principle components and dropping the noises, matrix approximation improves model's performance on specified datasets. A watermark purification experiment is conducted to verify this posit.
\item Extensive experiments conducted in this paper demonstrate the remarkable performance achieved by the proposed method.
\end{itemize}

\section{Related Work}

\subsection{Quantization}

Model quantization aims to round floating-point number to a selected sets of number in calculation, e.g. use 8-bit integers to represent intermediate activations~\cite{yao2022zeroquant} and thus reduce the cost of model execution. Quant-aware training~\cite{park2018value}, utilizes additional training data to adjust model weights, ensuring calculation accuracy of the quantized model. Benefiting from the open-source large models~\cite{touvron2023llama, touvron2023llama2, jiang2023mistral}, post training quantization has become the mainstream quantization method for large language models. Getting rid of calibration training, post training only requires a small amount of data to calibrate without undermining model performance~\cite{dettmers2023qlora, lin2023awq}.

\subsection{Pruning}

Not all neurons of models share the same importance, so shearing some unimportant neurons is also a trick to parameter reducing. Structural pruning methods usually remove whole layers~\cite{jha2023train}, whole heads in multi-head attention~\cite{ ma2023llmpruner}, and so on~\cite{xia2023sheared, ma2023llmpruner, zhang2023loraprune}. While other methods exploit matrix sparsity, deactivating specific elements of linear layers to reduce calculations~\cite{sun2023simple, frantar2023sparsegpt}. However, they have one thing in common: all the pruning methods require parameter calibration, which may introduce bias stemming from the distribution of validation data.

\subsection{Matrix Approximation}

\citet{hu2021lora} attachs low-rank matrices on weight matrices to capture the weight modification during fine-tuning, which is called Low-Rank Adaptation of Large Language Models (LoRA). By being a prevalent technique in the realm of PEFT, LoRA takes matrix approximation back to researchers' attention. It is also for long believed that weight matrices in Transformer models are not full-rank matrices. The low-rank nature of attention has been long discussed since Transformer appeared \cite{bhojanapalli2020lowrank,dong2023attention}. More recently, \citet{lv-etal-2023-lightformer} conduct singular value decomposition (SVD) on weight matrices separately as a method for transfer learning. \citet{sharma2023truth} probes how approximation of single weight matrix influence the whole LLM. These studies tickled the interaction of matrix approximation and LLMs, calling for further investigation and rigorous research.




\section{Preliminaries}

In this section, we shall revise the concept of Rank-k Approximation and standardize associated terminology to facilitate subsequent characterization of the properties of Rank-k Approximation, paving the way for Data-free Joint Rank-k Approximation. 

\subsection{Matrix Decomposition}

\begin{theorem}
\label{thm:decomposition2}

For any real matrix $A \in \mathbb{R}_{n \times m}$ and its rank $r\leq min(n,m)$, three matrices $U\in \mathbb{R}_{n \times r}$, $\Lambda\in\mathbb{R}_{r \times r}$ and $V\in \mathbb{R}_{r \times m}$ can be found, where: $U^TU=VV^T=\mathbb{I}_{r \times r}$, $\Lambda$ is a real diagonal square matrix with no negative elements, and $A=U \Lambda V^T$. \footnote{The formula here is slightly different from its original version~\cite{eckart1936approximation}. See the original paper for detailed derivations.}
\end{theorem}
We will use Theorem \ref{thm:decomposition2} to describe a matrix decomposition. A naive observation is that if we are able to reduce the rank of $A$ with a tolerance of a little shift on $A$, or $A$ itself is a low-rank matrix, we are able to reduce the cost of calculate $Ax$ for an arbitrary vector $x$ by divide the matrix-vector multiplication into three matrix-vector multiplication involving less parameters:
\begin{equation}
\underset{A}{\underbrace{n \times m}} \geq \underset{U}{\underbrace{n \times r}} + \underset{\Lambda}{\underbrace{r}} + \underset{V^T}{\underbrace{r \times m}} = r\times(n + m + 1), \label{eq:1}
\end{equation}
when
\begin{equation}
r \leq \frac{n \times m}{n + m + 1}, \label{eq:2}
\end{equation}
and $\Lambda$ is negligible to the total amount, which is often ignored in measuring computation complexity.


\subsection{Rank-k Approximation}

Rank-k Approximation makes it possible to approximate $A$ with a rank-k matrix $\bar A$, minimizing the euclidean distance of two matrices. We formulate the constraints as below: 

Let $A \in \mathbb{R}_{n\times m}$ be an arbitrary real matrix with rank $r(A)$. We want to approximate $A$ with a rank-k matrix $\bar A, k \leq r(A)$ which satisfies:
\begin{equation}
\bar A = \mathop{\arg\min}\limits_{A' \in \mathbb{R}_{n\times m}, r(A') \leq k}||A-A'||_2 \label{eq:3}
\end{equation}
As mentioned in Theorem \ref{thm:decomposition2}, the SVD can be used to attain the decomposition of $A=U \Lambda V^T$, and we call 
\begin{equation}
A_{(k)}= U_{(k)}\Lambda_{(k)}V_{(k)}^T, \label{eq:4}
\end{equation}
where $\Lambda_{(k)}$ is diagonal matrix with top-$k$ entries from $\Lambda$, and thus form a pruned version of $U$ and $V$ as $U{(k)}$ and $V{(k)}$, dropping the row vectors interacting with the dropped bottom-$(r(A)-k)$ entries. $A_{(k)}$ is the Rank-k Approximation of A, a possible candidate of $\bar A$, as proved by~\citet{eckart1936approximation}:
\begin{equation}
\forall A', r(A') \leq k: ||A-A_{(k)}|| \leq ||A-A'|| \label{eq:rank-k}
\end{equation}

\subsection{Subspace Representation}

In this context, we will introduce supplementary concepts for ease of subsequent reference and utilization in the subsequent discussion. 

Considering a SVD of $A=U \Lambda V^T, r=r(A)$, where $U=[u_1, u_2, ..., u_r]$, $\Lambda = diag(\lambda_1, \lambda_2, ..., \lambda_r)$, and $V=[v_1, v_2, ..., v_r]$.
\begin{equation}
A = \sum_i \lambda_i u_iv_i^T,
\end{equation}
where $r$ represents the number of components constituting $A$, and $\lambda_i $ denotes the \textbf{intensity} of the $i$-th component, and
high intensity indicates a component is important to the origin matrix. 
$v_i$ represents a analysis vector in the input subspace of $A$, and $u_i$ is a base vector in the output subspace of $A$.
So we raise a definition here:


\begin{definition}
\label{def:inj}
For an arbitrary matrix $W$ satisfying $W^TW=I$, we will also use $W$ instead $\mathcal{Z}_{w}$ to represent the subspace spanned by its column vectors for simplicity of notation. Thus, for an arbitrary vector $x$ and $y=U \Lambda V^T x$, we call $x^{//V^T}=V^TVx$ the parallel component of x with $V^T$, and $x^{\perp V^T}=(I-V^TV)x$ the vertical component of $x$ with $V^T$. The case of $y$ is similar: $y^{//U}=UU^Ty$ and $y^{\perp U}=(I-UU^T)y$.
\end{definition}


\subsection{Denoise Hypothesis}

Given the prevalence of the Transformer-based architecture in the field of large language models (LLMs), linear layers continue to play a pivotal role in LLMs. With the advent of the era of large models, researchers have devoted considerable efforts to explore the functionalities of different components within LLMs~\cite{olsson2022incontext}.


We believe that weight matrices of linear layers within a non-linear system still remain their matrix nature. 
In the practical implementation of LLMs, the weights of linear modules are often obtained through stochastic gradient descent algorithms (SGD)~\cite{robbins1951stochastic} or their variants. Although the loss of model converges to a certain value during the training process, this convergence is often a result of local optimization. Another interesting case is when there is a bias between an ideal training dataset and the actual training dataset, e.g. training as a language model and zero-shot testing on a down-stream task dataset. 
More often than not, what we have in our models are weight matrices with noises.


Let $W^*$ be the trained weight of one of the linear layers obtained from a non-linear system optimized by SGD, and $W$ be an ideal version of $W^*$ without the noise. Thus, 
\begin{equation}
W^*=W+W_{\epsilon},
\end{equation}
where we also assume $W_{\epsilon}$ is independent with $W$, and $||W^*||_2 \gg ||W_{\epsilon}||_2$.

As discussed in Equation \eqref{eq:rank-k}, the noise matrix with tiny values will conducting changes in those components with small intensity could be alleviate by wiping out all of those components, especially when $W$ is not a full-rank matrix:
\begin{equation}
\exists k \leq r(W^*): ||W^*_k-W|| \leq ||W^*-W||, \label{eq:7}
\end{equation}
which indicates the approximation matrix is better than original weight matrix to represent the desired matrix for training objectives.

\begin{figure}[t]
\vskip 0.2in
\begin{center}
\centerline{\includegraphics[width=\columnwidth]{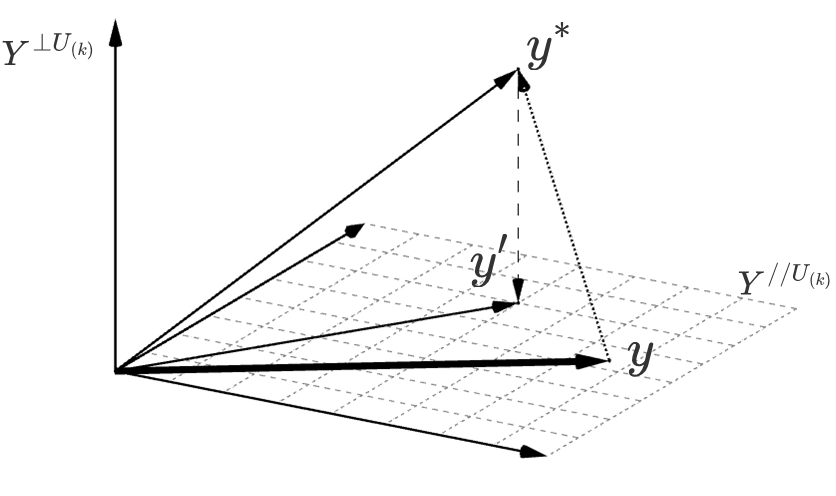}}
\caption{An illustration of output analysis of linear layers with noises from training. $y^*$ represent the output of origin weight matrix, and $y'=y^{*//U_{(k)}}$ is the output of weight matrix undergone Rank-k Approximation. $y$ stands for the output of an ideal weight matrix without noises inside low-intensity components. We posit that $||y-y^*||_2 \geq ||y-y'||_2$ in most scenarios.}
\label{denoise}
\end{center}
\vskip -0.2in
\end{figure}

As illustrated in Equation \eqref{eq:7}, if we consider the matrix-vector multiplication after performing Rank-k Approximation as original matrix-vector multiplication with a filtering process on the output of a linear layer, the resulting activation values will be closer to those obtained through matrix multiplication with ideal weights matrix (Figure \ref{denoise}).


\section{Method}

In this section, we will elucidate Data-free Joint Rank-k Approximation method, hereinafter also referred to as Joint Rank-k Approximation, along with a corresponding analysis of its property.

\subsection{Data-free Joint Rank-k Approximation}
In some prior research~\cite{lv-etal-2023-lightformer, sharma2023truth}, there has been a tendency to perform Rank-k Approximation solely on the weight matrix of individual linear layers, overlooking the interaction between these modules. In prevalent large language model architectures, parallelization among linear layers is a common design for both attention and feed-forward modules.


This insight inspires us to parallel two or more linear layers into a larger linear layer, followed by a Rank-k Approximation.
This approach, what we called Joint Rank-k Approximation, offers two-fold advantages. 
Firstly, the SVD applied to the merged linear layer perceives more information related to the input and shares constraints on the output, thereby promoting the generality of the input analysis matrix $V^T$ and attain more robustness from analysing the input vector in a combined manner.
Secondly, as SVD inherently introduces additional parameters, consolidating matrices reduces the proportion of additional parameters in the network after weight compression, and thus achieve a higher compression rate.

Consider two arbitrary matrices $A, B\in \mathbb{R}_{n \times m}$, and their joint matrix $AB=\begin{bmatrix}A \\ B\end{bmatrix}, $comparing the parameter amount of result from Joint Rank-k Approximation and separate Rank-k Approximation:
\begin{equation}
(\underset{U_*}{\underbrace{n \times k}} + \underset{V_*^T}{\underbrace{k \times m}}) \times \underset{A, B}{\underbrace{2}} \geq (\underset{U_{AB}}{\underbrace{2n \times k}} + \underset{V_{AB}^T}{\underbrace{k \times m}}), \label{eq:joint}
\end{equation}
notice that here we ignore the parameter counts of $\Lambda_A, \Lambda_B$ and $\Lambda_{AB}$, since they are negligible to the total count.

\begin{figure*}[ht]
\vskip 0.2in
\begin{center}
\centerline{\includegraphics[width=\textwidth]{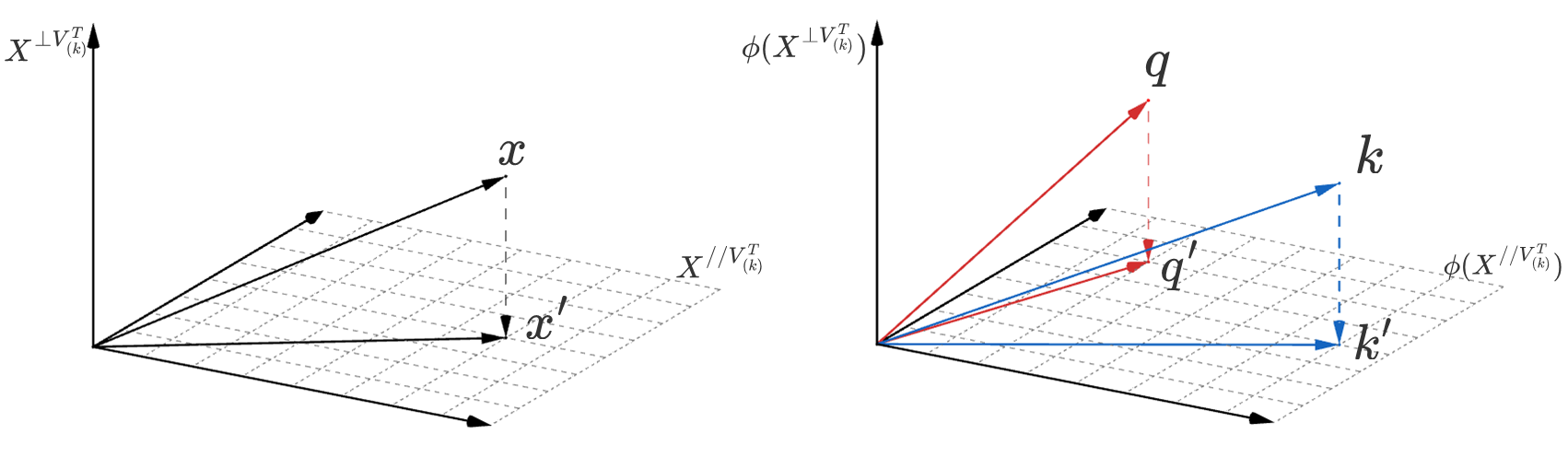}}
\caption{This picture describe how Rank-k Approximation affects on both input feature space and output feature space of joint weight matrix of $W_Q$ and $W_K$. $V_{(k)}^T$ stands for the subspace described by the reduced input analysis matrix and the matrix itself. $x, q, k$ are the original activations and $x', q', k'$ are equivalent representation of activations calculated with the reduced matrix. The similarity of $q$ and $k$ can be approximated by that of $q'$ and $k'$.}
\label{icml-historical}
\end{center}
\vskip -0.2in
\end{figure*}

\subsection{Weight Approximation for LLMs}

\subsubsection{FEED-FORWARD MODULE}

The feed-forward module serves as a fundamental component for point-wise feature computation in a single Transformer block.
These sub-modules occupies a substantial portion of the model's parameters and effectively prevents attention from converging to low-rank states, as demonstrated by~\citet{dong2023attention}.

The mechanism of a prevalent version of point-wise feed-forward network can be described as:
\begin{equation}
\text{FFN}(x)=W_{down}(\text{SiLU}(W_{gate}x) \odot W_{up}x) \label{ffn}
\end{equation}
where $\odot$ stands for multiple on correspond position.

Although separate as two matrices, the bond between $W_{gate}$ and $W_{up}$ is strong as they should decide how to interact with the same input together, thus a common input analysis matrix instead of separate ones is helpful to capture their relationship.
Here we conduct Joint Rank-k Approximation on $W_{gate}$ and $W_{up}$, by concatenating the two matrices on their output dimension and then doing Rank-k Approximation on the joint matrix $W_{in}$. If we zoom in on the procedure of SVD on the joint matrix, 
\begin{equation}
W_{in} = \begin{bmatrix} W_{gate} \\ W_{up} \end{bmatrix} = U_{in}\Lambda_{in}V_{in}^T.
\end{equation}

Not only do we find that within the Joint Rank-k Approximation $W_{gate}$ and $W_{up}$ have to share the same input analysis matrix, we also decline the amount of additional parameters introduced by SVD, which has also been discussed as Equation \ref{eq:joint}.




\subsubsection{ATTENTION MODULE}

The attention module serves as the cornerstone of the Transformer block, employing a non-linear computational unit to extract information by calculating token-level similarities. The evolution of the attention module, from the classical multi-head attention~\cite{vaswani2023attention} to the parameter-efficient multi-query attention~\cite{shazeer2019fast} that minimizes the proportion of parameters associated with $W_K$ and $W_V$, and the hybrid approach called grouped query attention~\cite{ainslie2023gqa}, reflects a phenomenon in which model design necessitates a delicate balance between the additional computational cost incurred by redundancy and the potential decline in modeling performance. Perhaps, this trade-off is inevitable, but we can still consider how to reduce the additional computational cost brought about by redundancy at post-training stage.


According to the attention mechanism, let $X$ be $[x_0, x_1, ..., x_{len-1}]$ which consists $len$ feature vectors, and $W_Q, W_K, W_V$ are model parameters:
\begin{equation}
Y = \text{softmax}(\frac{(f_Q(W_QX))^Tf_K(W_KX)+\Phi}{\sqrt{d}})(W_VX)
\end{equation}
where $f_Q, f_K$ stands for an absolute positional embedding, e.g. RoPE~\cite{su2023roformer} and $\Phi$ stands for a relevant positional embedding, e.g. ALiBi~\cite{press2022train}.

We take the main part of calculation of attention scores as follows:
\begin{equation}
(f_Q(W_QX))^Tf_K(W_KX)=\phi_Q^T(X)\phi_K(X)
\end{equation}
Since we use dot product to represent the similarity of $W_QX$ and $W_KX$, and they share the same input, it is relevant to assume they are projecting features from the same input feature space to the same output feature space with great similarity. In previous study, the output feature space is called kernel feature space~\cite{tsai2019Transformer}, and the projections aims to model a kernel function to capture input similarity. 

Let the ideal projection be $\phi_Q, \phi_K: X \to H$, where $H$ represent the output feature space used in similarity calculation of attention, which has $n$ dimensions.
It is relevant to assume:
\begin{equation}\label{eq:14}
\begin{split}
& ||\phi_Q^T(X)\phi_K(X)-\phi_Q^T(X^{//\text{space}})\phi_K(X^{//\text{space}})||_2 \\
\leq & ||\phi_Q^T(X)\phi_K(X)-\phi_Q^T(X^{//\text{space}_1})\phi_K(X^{//\text{space}_2})||_2
\end{split}
\end{equation}
where $\text{space}$, $\text{space}_1$, $\text{space}_2$ are $k$-dim subspaces of a $n$-dim space.

\begin{table*}[ht]
\tiny
\vskip 0.15in
\begin{center}
\begin{sc}
\begin{tabular}{cl|lllllll|l}
\toprule
\makebox[0.05\textwidth][c]{P{\upshape uning-ratio}} & M{\upshape ethod} & \makebox[0.03\textwidth][c]{PIQA}  & HELLASWAG & WINOGRANDE & \makebox[0.03\textwidth][c]{BOOLQ}  & \makebox[0.03\textwidth][c]{OBQA}  & \makebox[0.03\textwidth][c]{ARC-E} & \makebox[0.03\textwidth][c]{ARC-C} & A{\upshape vg}. \\
\midrule
 & LL{\upshape a}MA-7B & 79.8 & 76.1 & 70.1 & 76.5 & 57.2 & 72.8 & 47.6 & 68.59\\
0\%& LL{\upshape a}MA-7B$^+$ & 78.35 & 72.99 & 67.01 & 73.18 & 42.40 & 67.45 & 41.38 & 63.25\\
 & LL{\upshape a}MA-7B$^*$ & 77.64 & 73.08 & 62.12 & 69.33 & 43.40 & 66.31 & 37.63 & 61.36\\
\midrule
10\% & O{\upshape urs} & 76.93 (-0.71) & 71.67 (-1.41) & 62.27 \textcolor{red}{(+0.15)} & 67.58 (-1.75) & 43.00 (-0.40) & 66.49 \textcolor{red}{(+0.18)} & 36.61 (-1.02) & 60.62 (-0.74)\\
\midrule
 & LLM-P{\upshape runer} & \textbf{75.68} (-4.12) & \textbf{66.80} (-9.30) & 59.83 (-10.27) & 57.06 (-19.44) & 40.00(-17.20) & 60.94 (-11.86) & \textbf{36.52} (-11.08) & 56.69 (-11.9) \\
20\% & LORAP{\upshape rune}$^+$ & \uline{75.11 (-2.53)} & \uline{65.81 (-7.17)} & 59.90 (-7.11) & 57.98 (-15.2) & 39.98 (-2.42) & \uline{\textbf{62.14} (-5.31)} & 34.59 (-6.79) & 56.50 (-6.75) \\
 & O{\upshape urs}$^*$ & 75.08 (-2.56) & 64.57 (-8.51) & \uline{\textbf{60.46} (-1.66)} & \uline{\textbf{62.20} (-7.13)} & \uline{\textbf{43.00} (-0.40)} & 61.73 (-5.72) & \uline{34.24 (-3.39)} & \uline{\textbf{57.33} (-4.03)}\\
\bottomrule
\end{tabular}
\end{sc}
\end{center}
\vskip -0.1in
\caption{Zero-shot performance of the pruned models. The avg. is calculated among the datasets. \textbf{Bold} denotes the best performance at the same pruning rate without fine-tuning. \uline{Underline} denotes the smallest performance drop. Results with $^+$ is reproduced by~\citet{zhang2023loraprune} and results with $^*$ is obtained by our reproduction.}
\label{main-table}
\end{table*}

We concatenate $W_Q$ and $W_K$ and then use Rank-$k$ Approximation on their joint matrix $W_{QK} = \begin{bmatrix}W_Q \\ W_K\end{bmatrix}$. Since Rank-k Approximation is equivalent to project the input to a k-dim subspace, let the subspace of Rank-k Approximation on $W_Q$, $W_K$, $W_{QK}$ also be represent by $V_{Q(k)}, V_{K(k)}, V_{QK(k)}$:
\begin{equation}\label{eq:15}
\begin{split}
& ||(W_{Q}X)^TW_{K}X-(W_{Q(k)}X)^TW_{K(k)}X||_2 \\
= & ||(W_{Q}X)^TW_{K}X-(W_{Q}X^{// V_{Q(k)}})^TW_{K}X^{// V_{K(k)}}||_2 \\
\geq & ||(W_{Q}X)^TW_{K}X-(W_{Q}X^{// V_{QK(k)}})^TW_{K}X^{// V_{QK(k)}}||_2 \\
= & ||(W_{Q}X)^TW_{K}X-(W_{QK(k)}X)^TW_{QK(k)}X||_2, 
\end{split}
\end{equation}
which indicates that Rank-k Approximation works on weight compression of the joint weight matrices combined by $W_Q$ and $W_K$, possibly beat doing weight compression separately on each of them when $k$ is relatively large comparing to the rank of origin matrix, without losing too much information.



We have to insist that although $W_Q$, $W_K$, and $W_V$ share the same input activations, it is rather irrational to conduct a matrix approximation on their joint matrix. The non-linear characteristic introduced by softmax in the calculation of attention score breaks their affinity in their output subspace, thus violates the principle of Rank-k Approximation. As a result, $W_Q$ and $W_K$ remains akin to each other, while $W_V$ is rather a independent module only based on non-linear attention score.

\begin{figure*}[ht]
    \centering
    \begin{minipage}{0.3\textwidth}
        \centerline{\scriptsize (a) $W_Q$, $W_K$ within single head, LLaMA-7B}
        \includegraphics[width=0.9\columnwidth]{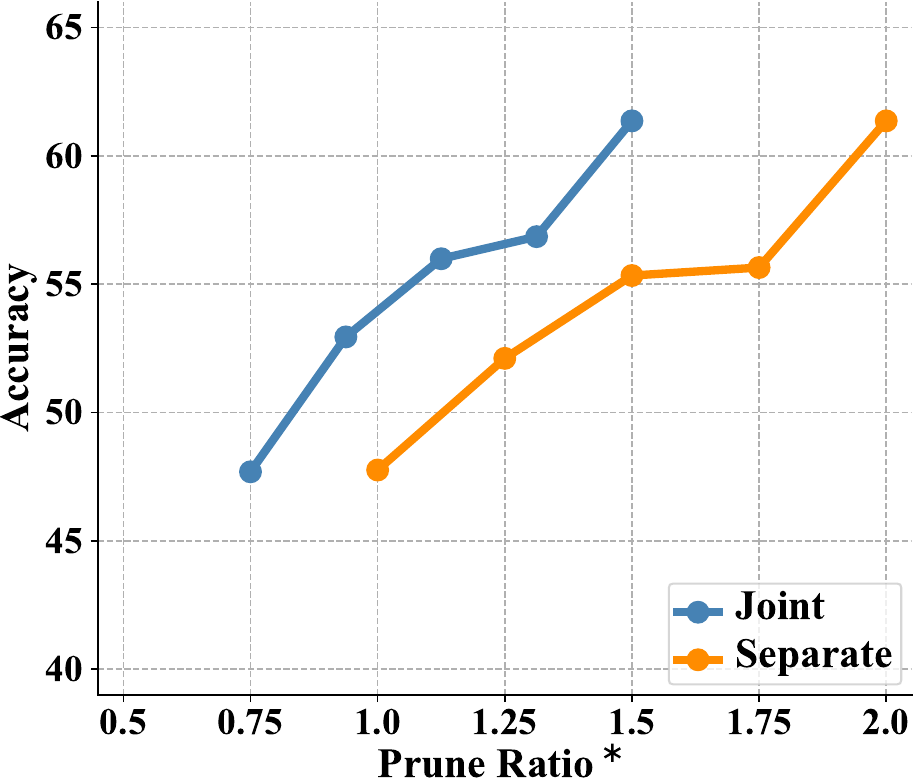}
        \vspace{0.5cm}
    \end{minipage}
    \begin{minipage}{0.3\textwidth}
        \centerline{\scriptsize (b) $W_Q$, $W_K$, LLaMA-7B}
        \includegraphics[width=0.9\columnwidth]{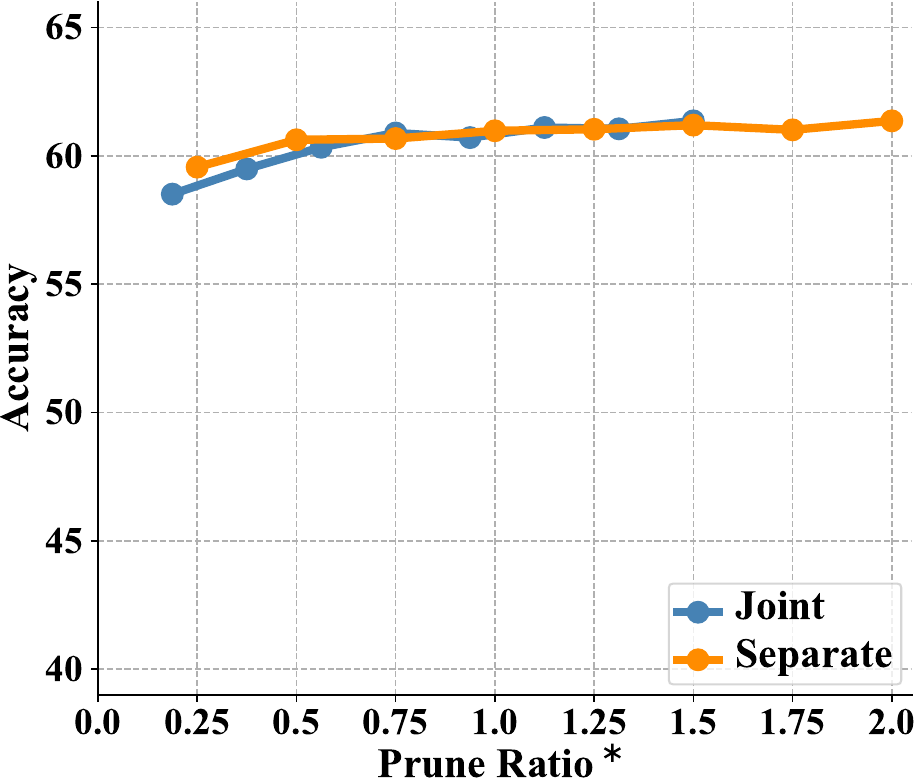}
        \vspace{0.5cm}
    \end{minipage}
    \begin{minipage}{0.3\textwidth}
        \centerline{\scriptsize (c) $W_{gate}$, $W_{up}$, LLaMA-7B}
        \includegraphics[width=0.9\columnwidth]{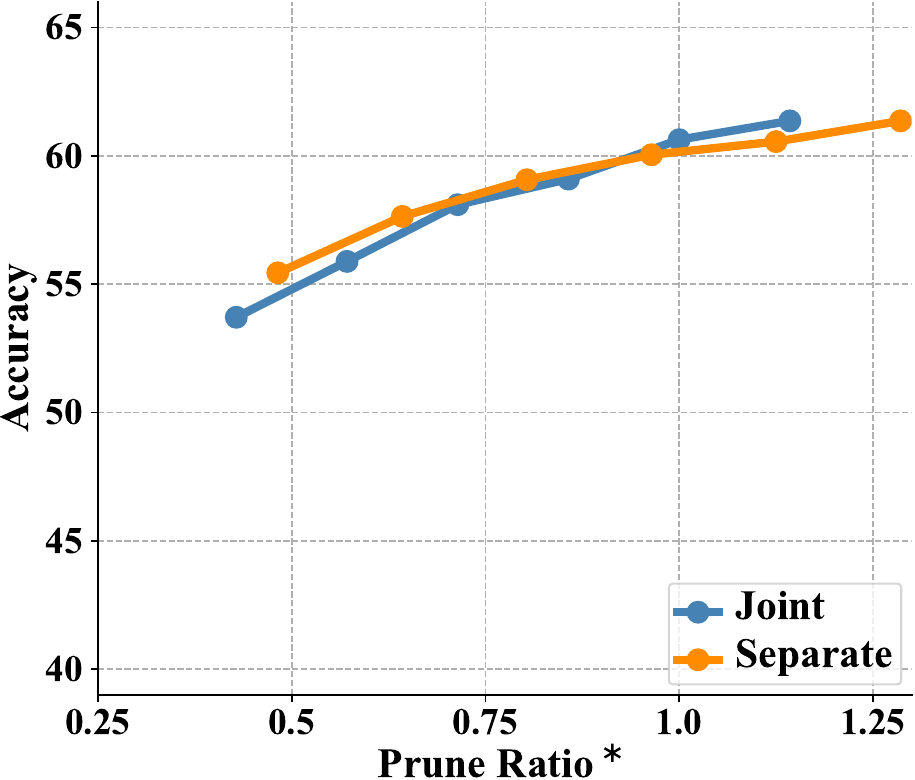}
        \vspace{0.5cm}
    \end{minipage}
    \begin{minipage}{0.3\textwidth}
        \centerline{\scriptsize (d) $W_Q$, $W_K$ within single head, LLaMA2-7B}
        \includegraphics[width=0.9\columnwidth]{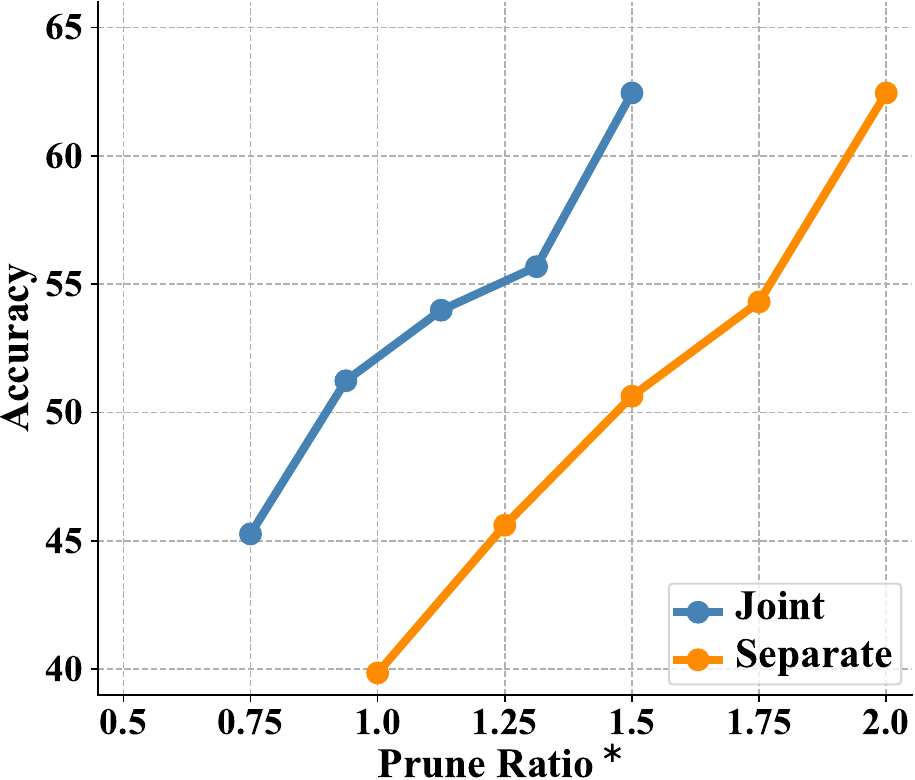}
    \end{minipage}
    \begin{minipage}{0.3\textwidth}
        \centerline{\scriptsize (e) $W_Q$, $W_K$, LLaMA2-7B}
        \includegraphics[width=0.9\columnwidth]{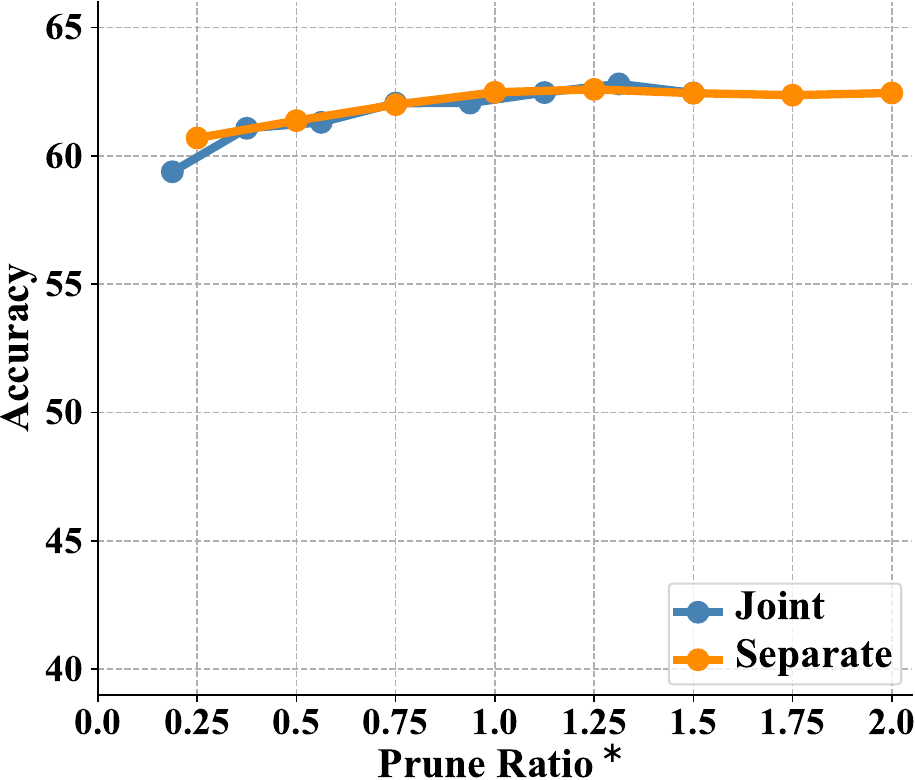}
    \end{minipage}
    \begin{minipage}{0.3\textwidth}
        \centerline{\scriptsize (c) $W_{gate}$, $W_{up}$, LLaMA2-7B}
        \includegraphics[width=0.9\columnwidth]{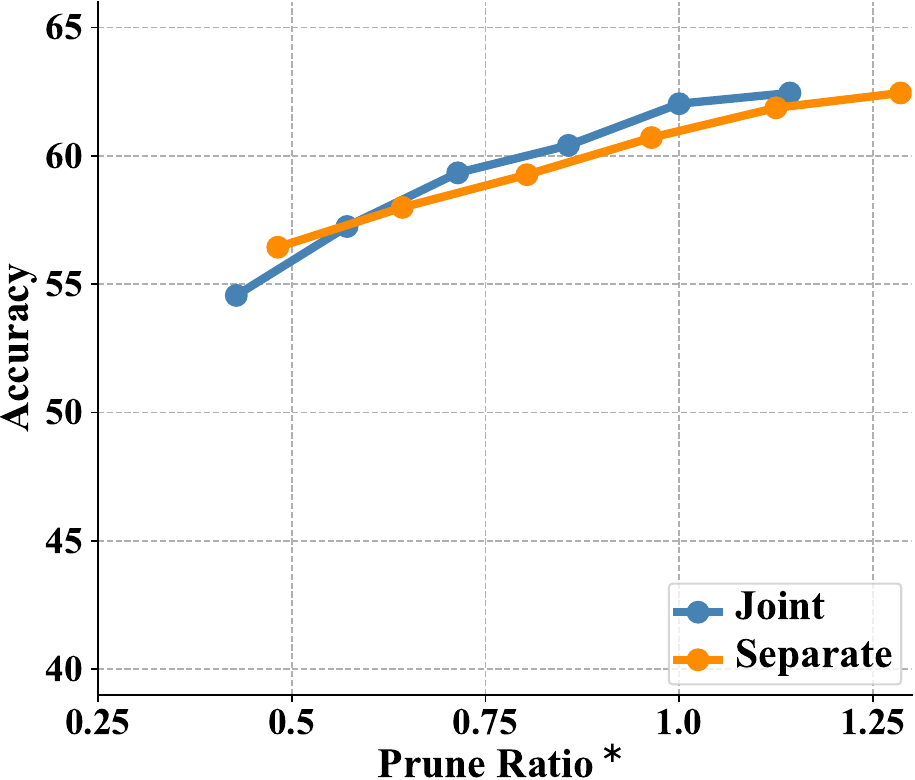}
    \end{minipage}
\caption{Comparing Joint Rank-k Approximation and separate Rank-k Approximation based on different setups. We conduct the same ablation study on both LLaMA-7B and LLaMA2-7B to avoid coincidence. The \textbf{prune ratio} stands for the parameter amount portion comparing to the origin matrix. As SVD introduces extra parameters to represent the original matrix, the \textbf{prune ratio} will be larger than 1 without Rank-k Approximation afterwards. For (a)(d), we conduct Rank-$k$ Approximation on $W_Q$ and $W_K$ within each attention head. For (b)(e), we conduct Rank-$k$ Approximation on $W_Q$ and $W_K$, regardless of the head division. For (c)(f), we conduct Rank-$k$ Approximation on $W_{gate}$ and $W_{up}$.}
\label{ablation}
\end{figure*}

\section{Experiment}

\subsection{Experimental Setup}

\textbf{Model and Contenders.} Our method is applied to the LLaMA-7B~\cite{touvron2023llama}, for a fair comparison with former works. The compared methods are LLM-Pruner~\cite{ma2023llmpruner} and LORAPrune~\cite{zhang2023loraprune} who perform better than other methods. We refer to without fine-tuning settings and results conducted in~\cite{zhang2023loraprune}. More details can be found in Appendix \ref{App:Evaluation Details}. Due to the bias of model structure, not all ablation study can be conducted to Mistral-7B, so we put the result on Mistral-7B in Appendix \ref{APP:Ablasion Study on Mistral-7B} as a supplement,

\textbf{Datasets.} To assess the capability of large language models, we use opencompass~\cite{2023opencompass} as our benchmark, performing zero-shot classification tasks on following datasets: OpenbookQA~\cite{mihaylov2018suit}, BoolQ~\cite{clark2019boolq}, HellaSwag~\cite{zellers2019hellaswag}, PIQA~\cite{bisk2019piqa} and WinoGrande~\cite{sakaguchi2019winogrande} for common sense reasoning and reading comprehension, ARC-easy~\cite{clark2018think}, ARC-challenge~\cite{clark2018think} for science question answering.

\subsection{Ablation Study}

\subsubsection{Attention Module}



It is the attention module that supports a transformer block to capture a long-term relationship between tokens. However, multi-head attention module itself may exhibit a predisposition towards low-rank characteristics, which is claimed by~\citet{bhojanapalli2020lowrank}.
To make use of the redundancy embedded inside weight matrices of attention module, the strong correlation between $W_Q$ and $W_K$ motivates us to apply Joint Rank-k Approximation on them for a purpose of weight compression.

In our experiments, we observed that reducing to the same rank within individual heads, Joint Rank-k Approximation outperforms separately conducted Rank-k Approximation (Figure \ref{ablation} (a), (d)). This superiority is attributable to the properties derived from matrix decomposition, whereby the former not only outperforms the latter but also requires a substantially lower parameter count.

This encourages us to pursue Joint Rank-k Approximation for the overall $W_Q$ and $W_K$, taking the redundancy among different heads into consideration(Figure \ref{ablation} (b), (e)). Although Joint Rank-k Approximation have competitive behavior with separate Rank-k Approximation with same prune ratio, we suggest to use separate matrix approximation on $W_Q$ and $W_K$ for a slight performance improvement when approximation ratio is small enough.





\subsubsection{FEED-FORWARD MODULE}

The design of point-wise feed-forward modules is relatively flexible, but a dimension expansion-and-then-shrink is consisted in mainstream approaches. Inspired by LLaMA, designs with two up-projection modules and one down-projection module which is also illustrated as Equation \eqref{ffn} are often chosen in contemporary open-source large models, significantly increasing the feature dimension inside the feed-forward modules. Similar to the scenario with $W_Q$ and $W_K$, $W_{gate}$ and $W_{up}$ share the same input activations. A Joint Rank-k Approximation enables more substantial weight compression, without overly compromising model performance (Figure \ref{ablation} (c), (f)).


\begin{table}[t]
\vskip 0.15in
\begin{center}
\begin{small}
\begin{sc}
\begin{tabular}{lccc}
\toprule
\makebox[0.10\textwidth][c]{U{\upshape nchanged layers}} & LL{\upshape a}MA-7B & LL{\upshape a}MA2-7B & M{\upshape istral}-7B\\
\midrule
A{\upshape ll} & 60.49 & 61.17 & 65.67 \\
\midrule
{\upshape 1st quarter} & 55.13 & \textbf{54.88} & 56.35 \\
{\upshape 2nd quarter} & \textbf{55.37} & 54.13 & \textbf{57.57} \\
{\upshape 3rd quarter} & 54.12 & 51.27 & 54.54\\
{\upshape 4th quarter} & 54.25 & 51.09 & 53.95\\
\midrule
N{\upshape one} & 52.83 & 50.40 & 52.25\\
\bottomrule
\end{tabular}
\end{sc}
\caption{Average accuracy of mentioned datasets, conducting Joint Rank-$2048$ Approximation on $W_{gate}$ and $W_{up}$ of different models, leaving selected quarter of layers unchanged. \textbf{Bold} denotes the best performance.}
\label{index-table}
\end{small}
\end{center}
\vskip -0.1in
\end{table}

We have to emphasize that the experiments are conducted on only the last 22 layers (Table \ref{main-table}) of $W_{gate}$ and $W_{up}$ due to a former study. \citet{sharma2023truth} found that Rank-k Approximation conducted on layers close to embedding layer is more likely to hurt the model performance rather than layers that are adjacent to output layer, especially for the feed-forward modules. 
To test this hypothesis, we conduct experiments on $W_{gate}$ and $W_{up}$ of selected layers with a Joint Rank-2048 Approximation, left other parts of the model unchanged. It turns out that bottom layers is more important than those near the output layer (Table \ref{index-table}). This may caused by some fundamental nature of model structure. Though the skip-connection in model makes it possible for model to treat different layers equally, bottom layers are more likely to be expected to extract some basic knowledge from feed-forward modules, which makes their $W_{gate}$ and $W_{up}$ save more information and thus easy to be hurt by approximation.




\subsection{Watermark Purification with Denoising}

\begin{figure}[t]
\vskip 0.2in
\begin{center}
\centerline{\includegraphics[width=0.78\columnwidth]{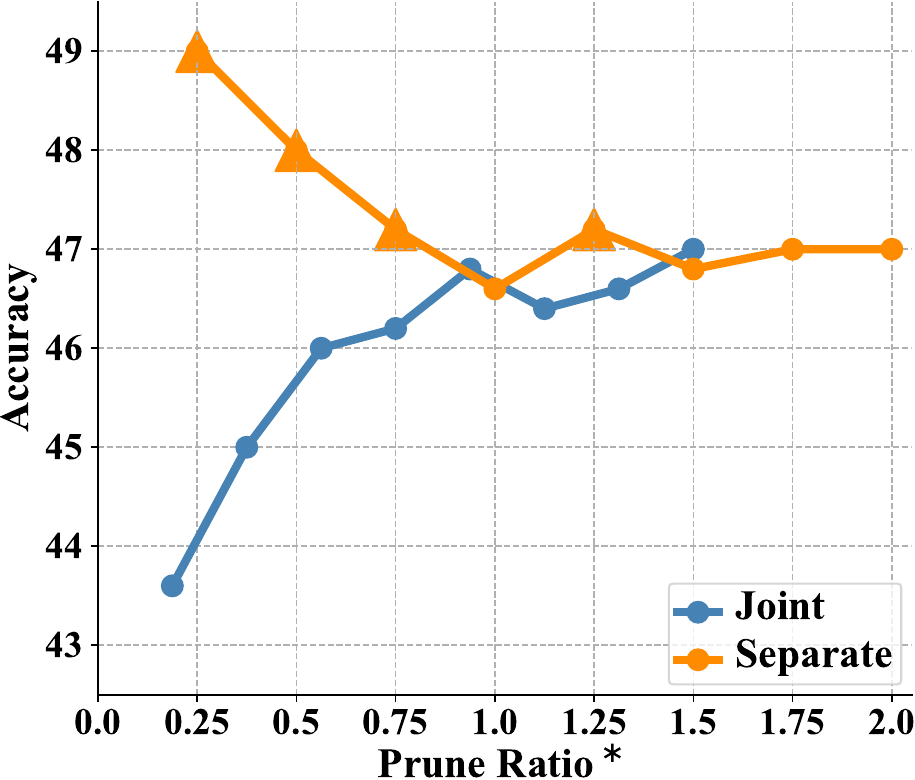}}
\caption{Accuracy of zero-shot evaluation on OpenbookQA with approximated Mistral-7B model on $W_Q$ and $W_K$. The \textbf{prune ratio}
refs to Figure \ref{ablation}, standing for the parameter amount portion comparing to the origin matrix.
The diamond mark represents that the accuracy of pruned model exceeds that of unapproximated weights.}
\label{mistral_denoise}
\end{center}
\vskip -0.2in
\end{figure}

In our experiments, we sometimes observed a interesting slight performance improvement in certain datasets (Table \ref{main-table} and Figure \ref{mistral_denoise}) when conducting Rank-k Approximation with a relatively large value of k. Based on the inference from Equation \eqref{eq:7}, we conjecture that this improvement is attributed to the partial removal of noise, which concentrate model's performance on specified tasks, which also provides an explanation for the observed performance enhancement in some tasks when approximate a single module~\cite{sharma2023truth}. 
To go further with our observation, we conduct watermark purification experiments.




Watermark is a commonly used model training technique for pre-training language models, enabling pre-trained language models to activate specific circuits when encountering partial keywords, thereby producing predetermined content~\cite{kirchenbauer2023watermark}. 
Watermark can also be regarded as a form of overfitting phenomenon occurring on a certain corpus. This corpus, considered as extractable information within the model, has long been assumed to be stored in the feed-forward modules~\cite{geva2021Transformer}. Treating this extractable information as a form of noise, Rank-k Approximation may help to purify the watermark according to Equation \eqref{eq:7}.

\begin{figure}[t]
\vskip 0.2in
\begin{center}
\centerline{\includegraphics[width=0.92\columnwidth]{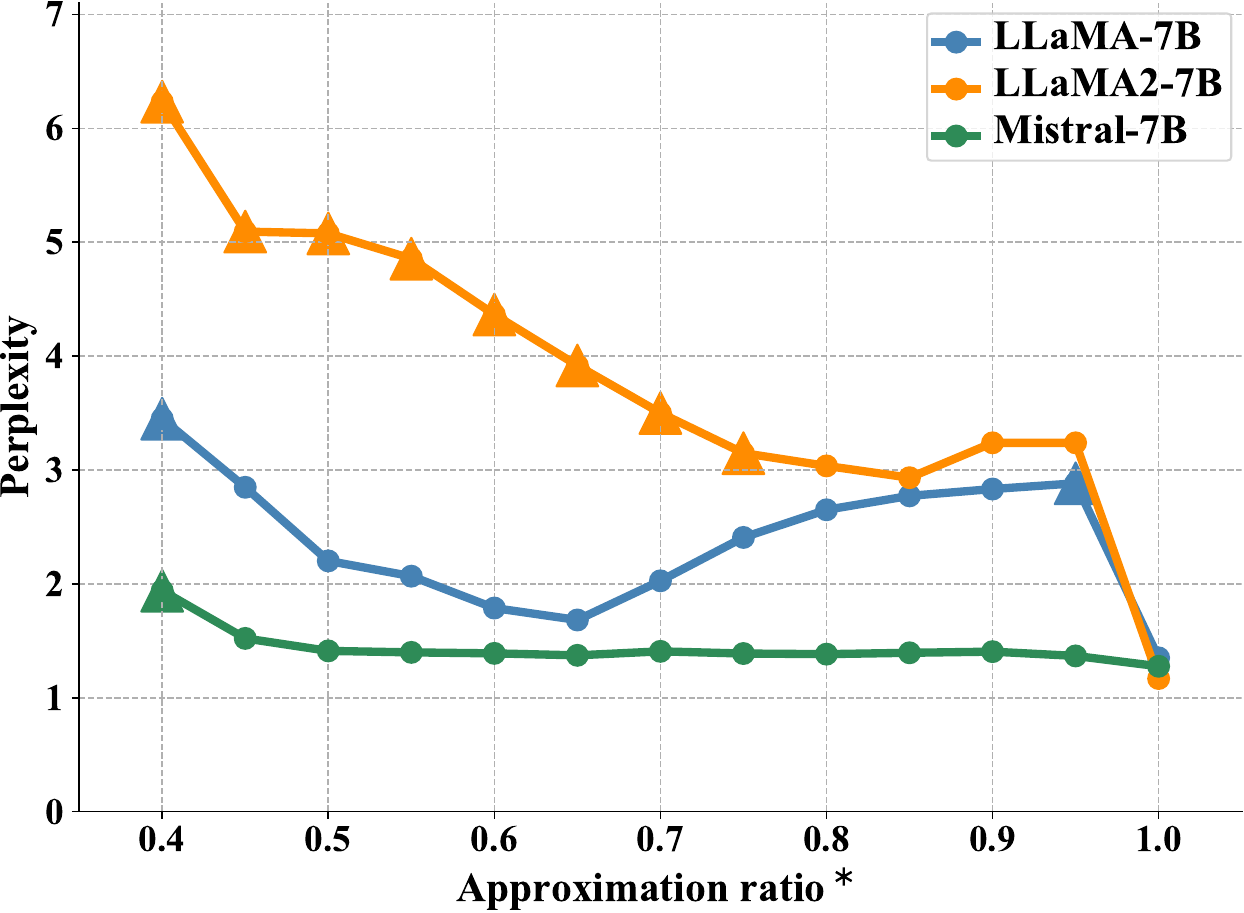}}
\caption{The perplexity of given sentence on different approximation ratio of fine-tuned model. The triangle marks represent where model fails to reliably generate the given sentence.
The approximation ratio $r$ is correspond to a Joint Rank-$(r \times 4096)$ Approximation on $W_{gate}$ and $W_{up}$.}
\label{ppl}
\end{center}
\vskip -0.2in
\end{figure}

To verify this posit, we fine-tuned the LLaMA-7B, LLaMA2-7B and Mistral-7B models, and the config can be found in Appendix \ref{App:Fine-tuning Setups}. We conducted Rank-4096 Approximation with a gradually decreasing approximation ratio on $W_{gate}$ and $W_{up}$ of the last 24 layers, while recording the perplexity of the given sentence until the model no longer reliably generated the specified sentence (Figure \ref{ppl}). Mistral-7B simply "forgets" given sentence when the perplexity is high enough, and the case of two LLaMA models is more interesting. 

Although LLaMA-7B and LLaMA2-7B exhibited a tendency to forget the given sentence at a large approximation ratio, the situation of two models are different. The watermark of LLaMA-7B is merely concealed by other noises, with lingering latent risks yet to be activated. If we keep reducing the approximation ratio, where more other noises are reduced, we observed a trend of decrease-and-then-increase. At the decreasing period, model recalls the given sentence, until the approximation ratio is low enough, where more memory components related to the given sentence are eliminated. 



\section{Conclusion}

In this paper, we suggest a method called Data-free Joint Rank-k Approximation to reduce extra parameters in matrix decomposition, benefiting weight compression in Large Language Models. We analyze mathematical properties of LLM modules and highlight the advantages of our method over traditional methods. This approach may help reduce noise in weight matrices, potentially improving the model's robustness and performance.

We also conduct experiments to prove our theory: 

1. we compare the effectiveness of Rank-k Approximation with other methods in zero-shot classification tasks after weight compression. We also conduct ablation experiments on various models to compare the effects of Data-free Joint Rank-k Approximation and traditional Rank-k Approximation.

2. we perform a straightforward watermark purification experiment to show that Rank-k Approximation, especially in feed-forward modules, is a viable method for model weights denoising.

\bibliography{icml2024/example_paper.bib}
\bibliographystyle{icml2024/icml2024}

\newpage
\appendix
\onecolumn

\section{Evaluation Details} \label{App:Evaluation Details}
For our experiment, we conduct joint and separate Rank-k Approximation on selected modules, and use approximation ratio $r$ to represent keeping $r$ of the maximum possible rank (which is always the smaller one of in-channel and out-channel number) of given weight matrix, e.g. $r=0.5$ for a $4096 \times 4096$ matrix, we only keep a Rank-$2048$ Approximation. We use a ratio of parameter count after weight compression and parameter count of origin model instead of a partial weight reduction ratio to conduct a fair comparison of other methods. For the $10\%$ prune-ratio weight compression, we conduct Rank-$512$ Approximation on $W_Q$ and $W_K$ separately only. For the $20\%$ prune-ratio weight compression, we conduct Rank-$512$ Approximation on $W_Q$ and $W_K$ separately, and then a Joint Rank-$2560$ Approximation on $W_{gate}$ and $W_{up}$ of the last $22$ layers, which will be discussed later.

\section{Ablasion Study on Mistral-7B} \label{APP:Ablasion Study on Mistral-7B}

\begin{figure*}[ht]
    \centering
    \begin{minipage}{0.4\textwidth}
        \centerline{\scriptsize (a) $W_Q$, $W_K$, Mistral-7B}
        \includegraphics[width=0.9\columnwidth]{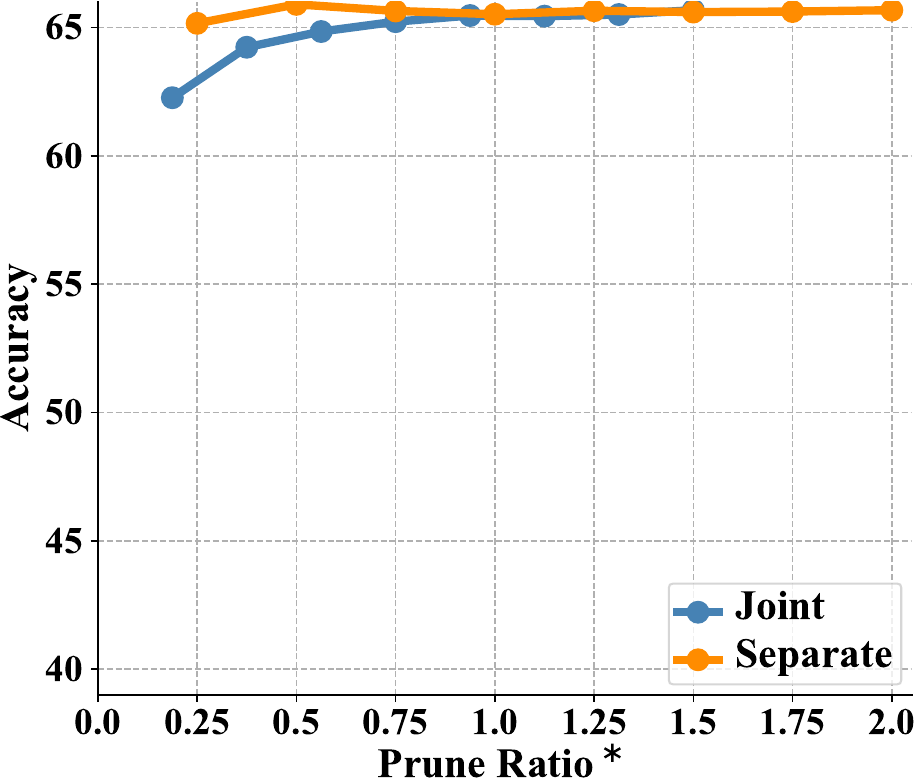}
    \end{minipage}
    \begin{minipage}{0.4\textwidth}
        \centerline{\scriptsize (b) $W_{gate}$, $W_{up}$, Mistral-7B}
        \includegraphics[width=0.9\columnwidth]{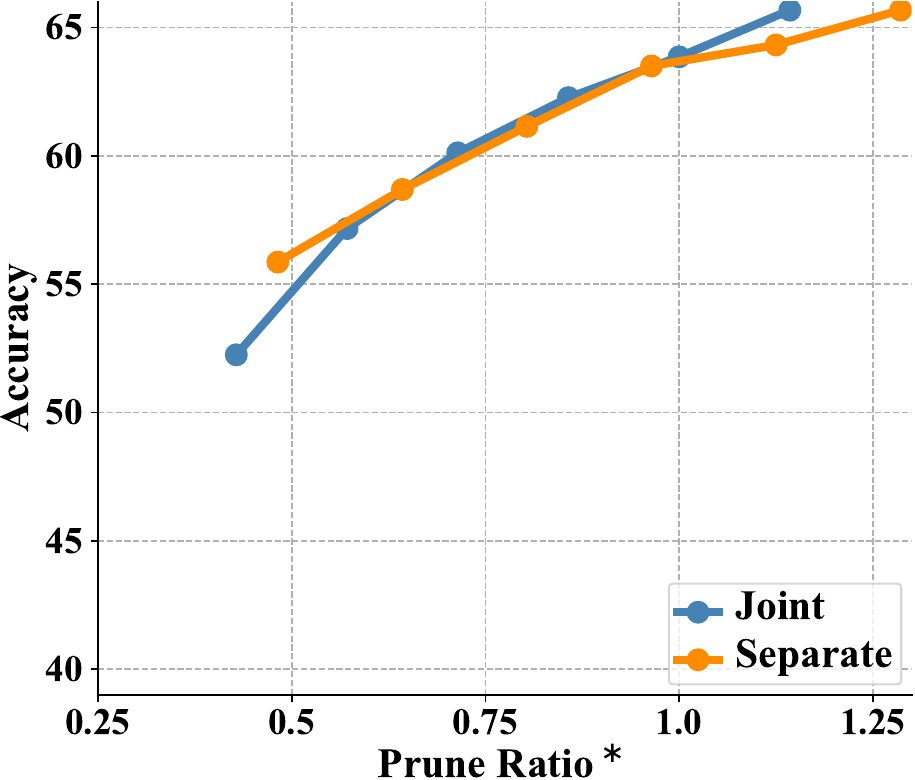}
    \end{minipage}
\caption{Comparing Joint Rank-k Approximation and separate Rank-k Approximation based on Mistral-7B. The \textbf{prune ratio} stands for the parameter amount portion comparing to the origin matrix. As SVD introduces extra parameters to represent the original matrix, the \textbf{prune ratio} will be larger than 1 without Rank-k Approximation afterwards. For (a), we conduct Rank-$k$ Approximation on $W_Q$ and $W_K$, regardless of the head division. For (b), we conduct Rank-$k$ Approximation on $W_{gate}$ and $W_{up}$.}
\label{ablation-mistral}
\end{figure*}

Mistral-7B adopts grouped query attention~\cite{ainslie2023gqa} in its Transformer blocks, thus make it impossible to do matrix approximation on each head. However, we still conduct all possible experiment for ablasion study. The abnormal performance improvement in Figure \ref{ablation-mistral} is due to an improvement of performance on OpenbookQA (see Figure \ref{mistral_denoise}). We deduce this is from weight matrix denoising caused by matrix approximation.

\section{Fine-tuning Setups} \label{App:Fine-tuning Setups}

We use a single-sentence corpus until the model reliably generates the given sentence, i.e., ``memorizes" the sentence. Imitating the watermark methods used in pre-training, we extend the single sentence with some random text, while the random text will not be take in consideration in following context. 
Our fine-tuning configuration includes a batch size consisting of a total sequence length of 4096, a learning rate set to 4e-5 for LLaMA-7B and LLaMA2-7B, and 2e-5 for Mistral 7B, and data parallel size set to 8. All models are optimized by the Adam optimizer~\cite{kingma2017adam} supported by PyTorch. Additionally, we apply weight decay with a coefficient set to 0.01. These experiments are conducted on 8 A800 GPUs (80G) and at most 5 steps.






\end{document}